\documentclass[12pt]{article}
\usepackage{epsfig,amssymb,amsmath}
\usepackage{alltt}
\usepackage{array}
\usepackage{latexsym}
\usepackage[english]{babel}
\usepackage{multicol}

\title{Logistic regression with MIL and $L_1$ regularization}
\author{Inna Stainvas}
\date{November 15, 2010}

\begin{document}
\title{Cancer Detection with Multiple Radiologists via Soft Multiple Instance Logistic Regression and $L_1$ Regularization}
\author{Inna Stainvas, Alexandra Manevitch, Isaac Leichter}

\maketitle

\begin{abstract}
This paper deals with the multiple annotation problem in medical application of cancer detection in
digital images. The main assumption is that though images are labeled by many experts,
the number of images read by the same expert is not large. Thus differing with the existing work on modeling
each expert and ground truth  simultaneously, the multi annotation information is used in a soft manner.
The multiple labels from different experts are used to estimate the probability of the findings to be marked as malignant.
The learning algorithm minimizes the Kullback Leibler (KL) divergence between the modeled probabilities and desired ones
constraining the model to be compact. The probabilities are modeled by logit regression and multiple instance learning
concept is used by us.

Experiments on a real-life computer aided diagnosis (CAD) problem for CXR CAD lung cancer detection demonstrate
that the proposed algorithm leads to similar results as learning with a binary RVMMIL classifier
or a mixture of binary RVMMIL models per annotator. However, this model achieves a smaller complexity and is more
preferable in practice.

\end{abstract}


\section{Introduction}
This paper deals with the multiple annotation problem in medical application of cancer detection in
digital images. In difference with the existing work on learning from multiple annotators \cite{Carpenter2008, raykar_ICML_2009_multiple_experts, YanRosales2010, WelinderNips2010},
in our application the same radiologist reads only a small subset of images from the entire database.
Our main assumption is that though the same image is read by many experts, the number of images read by the same expert is
not large. Thus instead of modeling each expert and ground truth separately; the multi annotation
information is used to deal with the noisy radiologist perception the way around.

The multiple radiologist marks are used to create a merged ground truth (GT) mark as an ellipse area (section~\ref{sec:gtMark}) and
then the ground truth (GT) marks are labeled in a soft manner by assigning them a probability to be malignant (section~\ref{sec:gtMarkProbability}).
A new probabilistic model with learning based on minimizing the Kullback Leibler (KL) divergence between the modeled probabilities and desired ones
and constraining the model's complexity is introduced by us. The probabilities are modeled by logit regression and multiple instance learning
concept is used by us (Section~\ref{sec:softMIL}).

\section{Multiple annotation}
\begin{figure}[htb]
\begin{center}
\begin{tabular}{ccc}
a & b & c\\
\includegraphics[height=3.5cm,width=2.8cm]{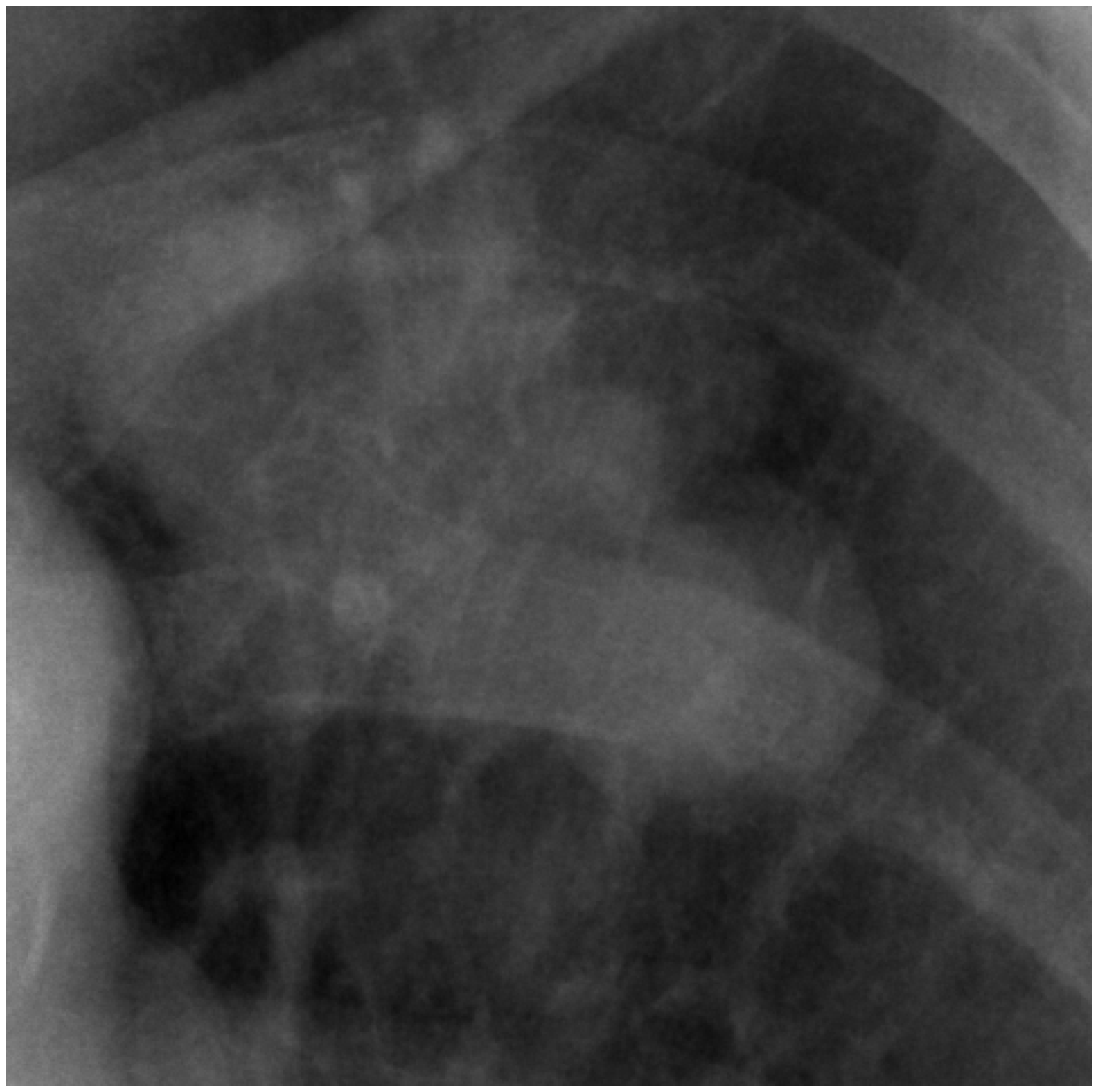} & \includegraphics[height=3.5cm,width=2.8cm]{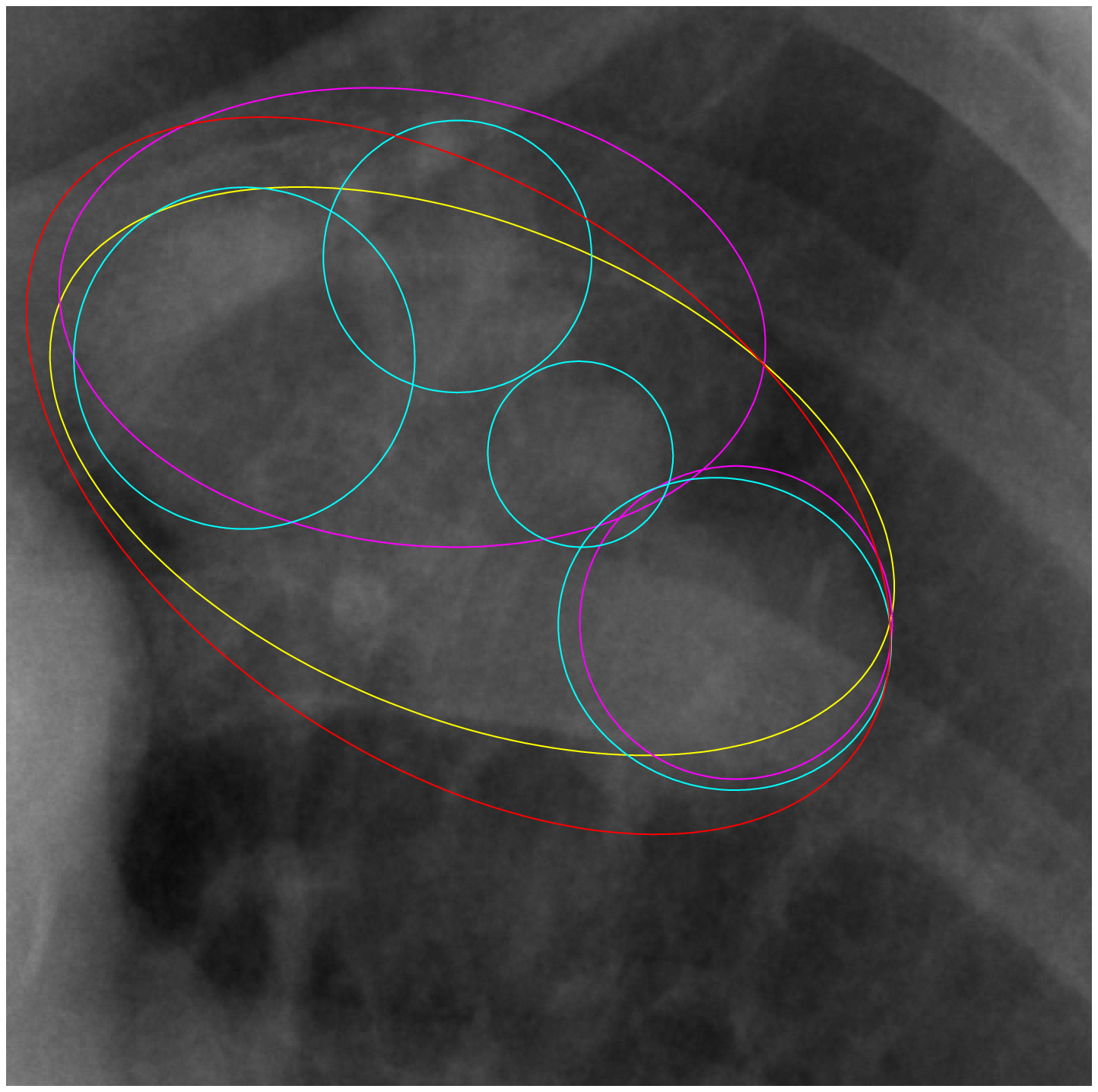}
& \includegraphics[height=3.5cm,width=2.8cm]{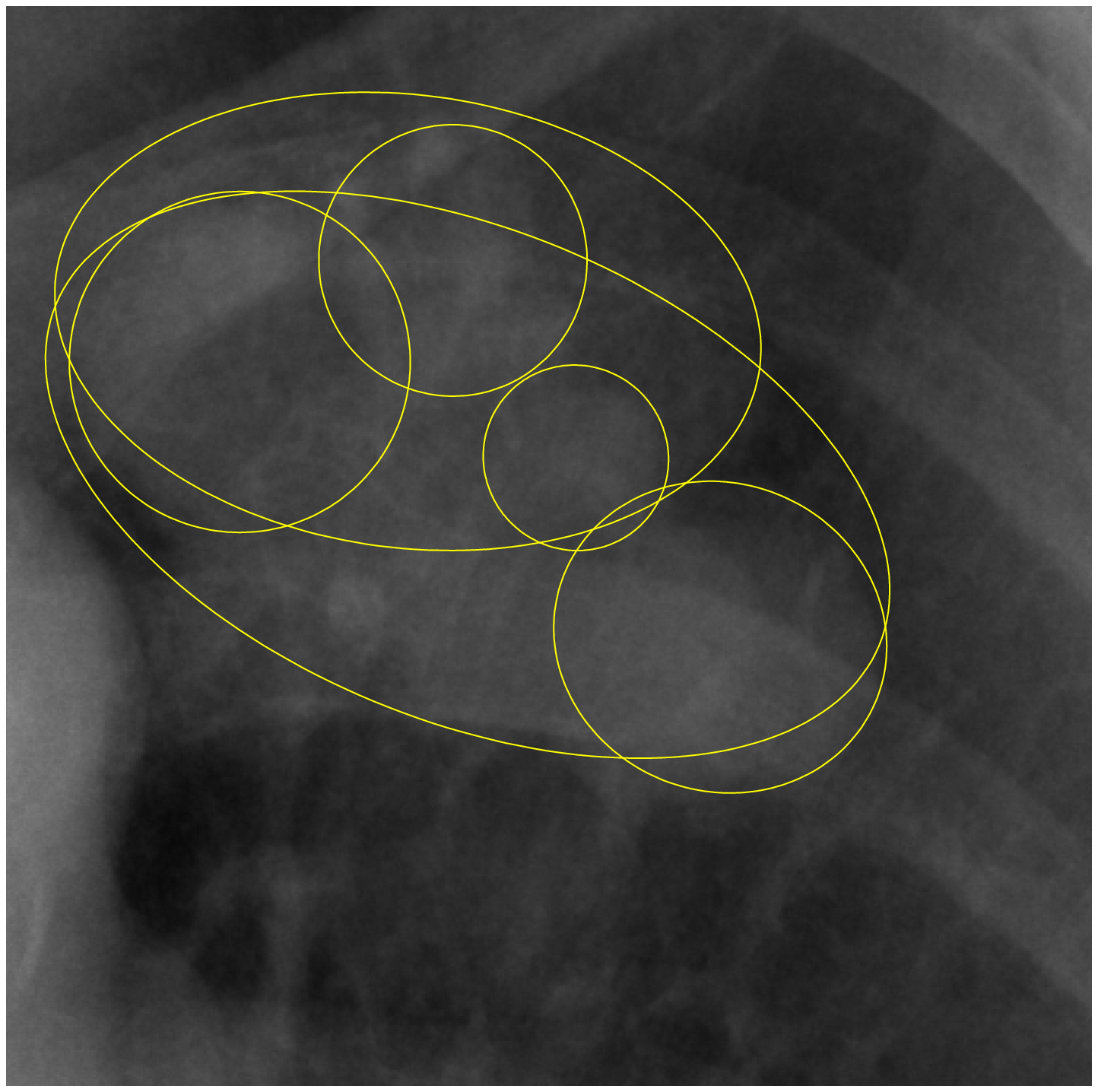} \\ \\
\includegraphics[height=3.5cm,width=2.8cm]{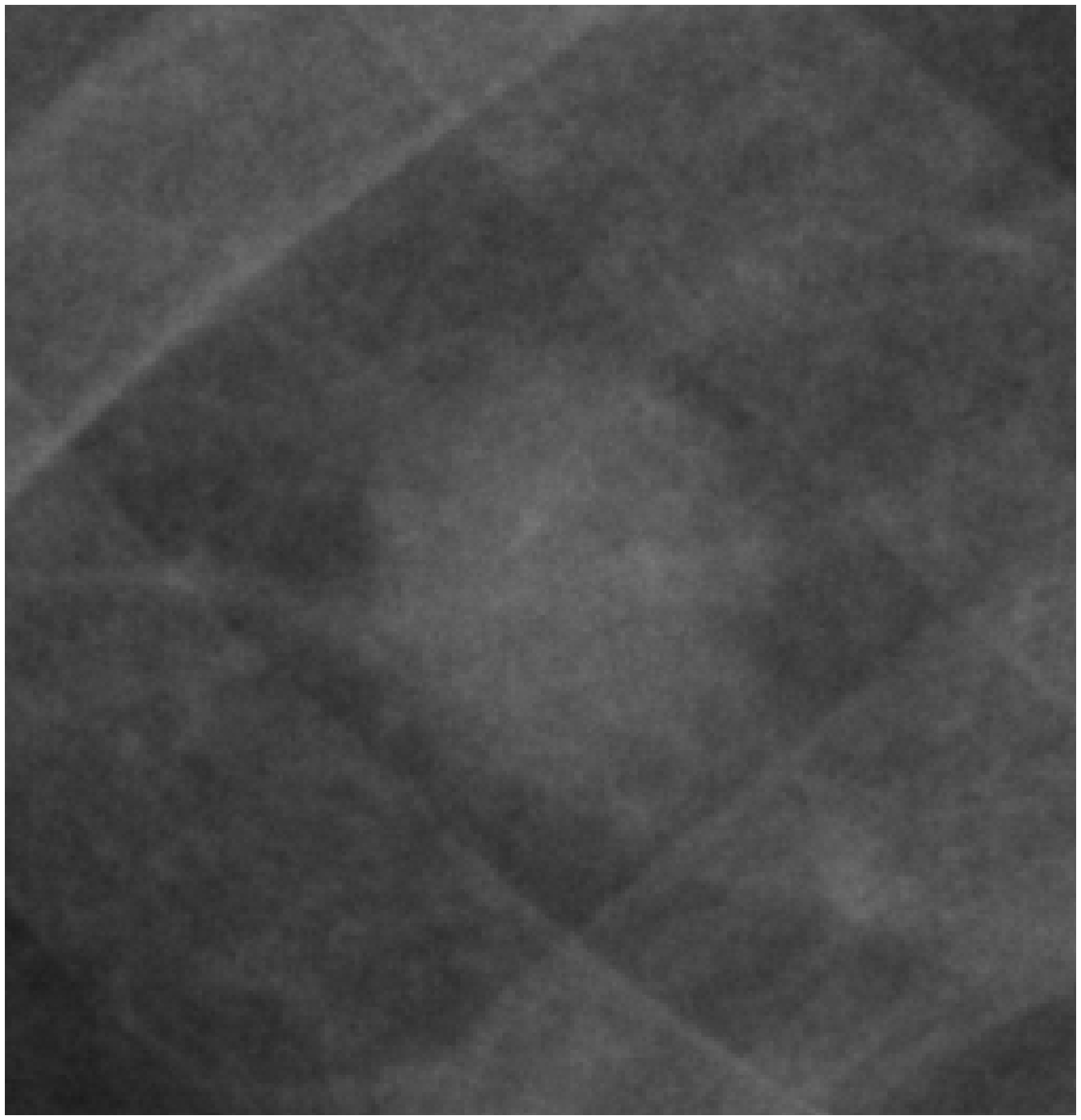} & \includegraphics[height=3.5cm,width=2.8cm]{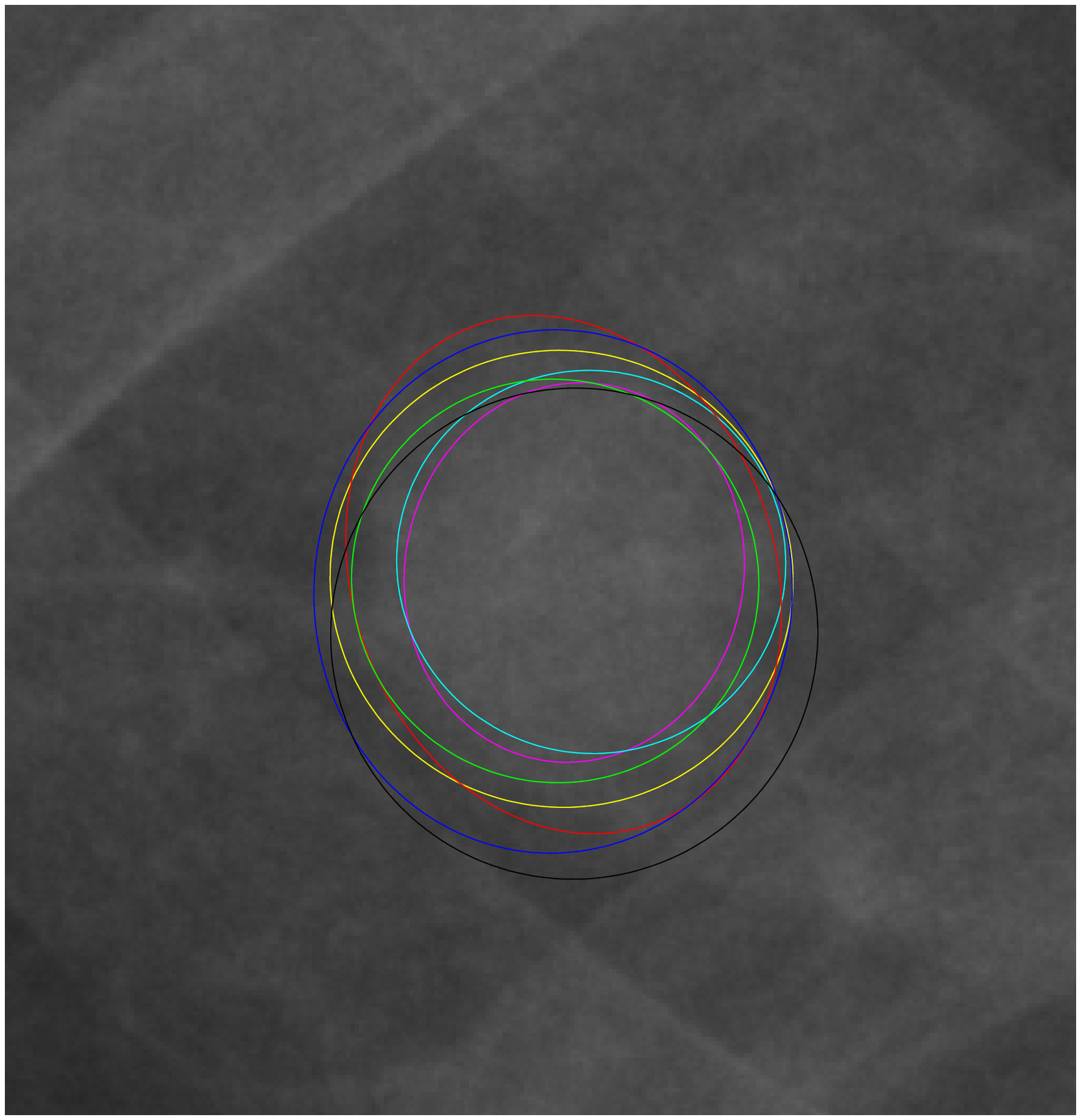}
& \includegraphics[height=3.5cm,width=2.8cm]{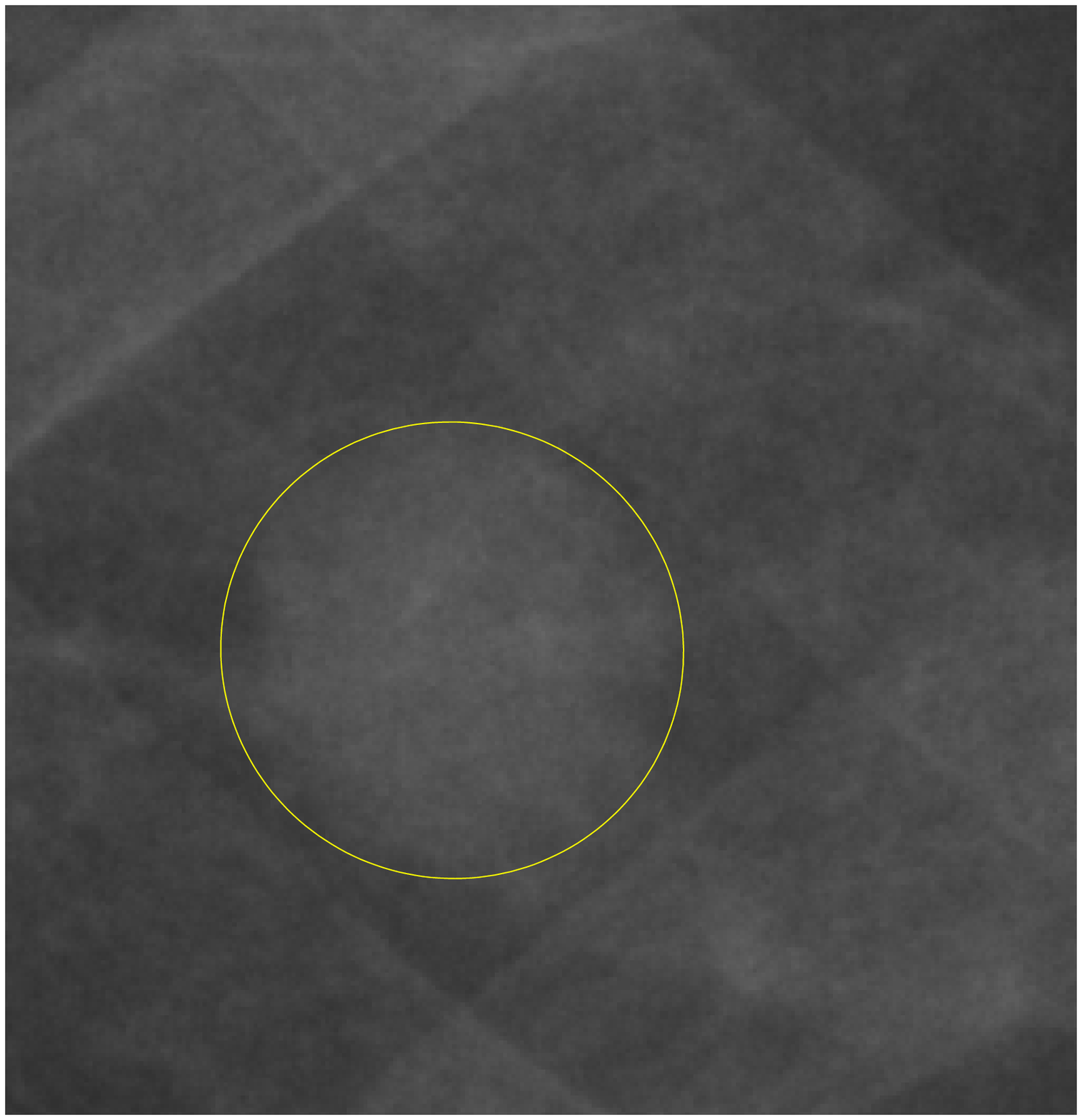}
\end{tabular}
\caption{{\sf Multiple annotation by ellipsoidal marks.} a$-$column: Pathological region displayed to radiologists.
b$-$column: The same region marked by several radiologists. Marks given by the same radiologist are drawn by the same color.
c$-$column: The final GT marks. The top row demonstrates a difficult case of different multi-scale perception by radiologists. The low row is
a simple case of the same scale perception but slightly different geometrical marking.}
\label{fig:MultipleAnnotation}
\end{center}
\end{figure}

It is well known that different radiologists perceive suspicious lesions differently.
When presented by the image and asking to mark the suspicious regions by ellipse they draw ellipses in a different position, orientation and scale.
Some radiologists have a tendency to draw a large ellipse covering the suspicious region while others prefer to present the same region
by drawing a few small ellipses (see Figure~\ref{fig:MultipleAnnotation}b, top row). Even for easy cases of a single perceptual scale their
marking may be still different, so that though the marking ellipses have a large area of intersection their parameters are still different
(see Figure~\ref{fig:MultipleAnnotation}b, bottom row).

In order to deal with this geometrical variability representing the same region, a ground truth ellipse (GT) is created by merging ellipses
drawn by all the annotators. In order to merge the marks of different annotators, they are split into groups representing the same GT
objects.

\subsection{Geometrical Merging} \label{sec:gtMark}
The merging GT is created by the following heuristic procedure.
\begin{enumerate}
\item First, all possible intersecting pairs of marks annotated by different radiologists are considered.
The marks are considered to be hitting (potentially referring to the same GT mark), if their normalized intersection area is sufficiently large
$\frac{S_{ij}}{max(S_i,S_j)} >T$,  where $S_i,~S_j,~S_{ij}$ are the marks and their intersection areas, respectively;
the threshold $T$ is selected adaptively to the mark's configuration.
When the marks are very similar in size\footnote{their size fraction is more than $0.7$} and position\footnote{distance between
mark centers normalized by their size is less than $0.1$}, the threshold is set to $T = \min(T_0, \frac{T_0}{20} \max(D_i,D_j))$; in all the other cases it is set to
$T = \min(T_0, \frac{T_0}{20} \min(D_i,D_j))$ and $T_0=0.63$, where $D_i,~D_j$ stand for the mark sizes in mm.
\item Iterate over the marks and assign to each mark a score equal to the number of marks which it hits.
Organize the marks into a list sorted by the score value.
\item {While the marks list is not empty, find the "seeding mark" of the largest score and merge it with all its hitting marks to create the primary GT mark.
In fact, the primary GT mark is an object referring to the list of marks consisting from the "seeding mark"
and marks hitting with it. The primary GT mark is represented as the mark of the median size in the list and has
a score equal to the number of marks in the list.
The exclusion from this is a case of the GTs referring to one or two marks only.
If the primary GT refers to two marks only, it is represented by one of them randomly selected.
The single marks are just stand for the primary GT of the score equal to one.} \label{item.0}
\item Marks already merged to the primary GTs are thrown from the mark list and do not participate in the subsequent merging process.
\item Continue this procedure iteratively from Step~\ref{item.0}
\end{enumerate}
When the procedure above is finished, the coinciding primary GTs are merged to final ones with the score being corrected
to indicate all the marks it refers to and to indicate their number as a final GT score. The non-coinciding primary GTs
are just the final GTs used in classification.

\section{Soft label creation} \label{sec:gtMarkProbability}
In our application we are not provided with the ground-truth labels obtained from biopsy or from other modalities;
but instead by many marks from different radiologists.
The geometrical aspects of GT marks generation from the multiple annotator responses were explained in section~\ref{sec:gtMark}.
However, the GT marks have additional score value that shows how many radiologists mark this GT object (consider this region as suspicious).

The number of radiologists is an important characteristic that enables to estimate the probability of the
GT to be perceived malignant. If all radiologists are of the same competence, the naive calculation by
the proportion of the positive (malignant) radiologist responses to the overall number of radiologists reading the image
is a good estimation of the probability of the mark to be malignant as perceived by the radiologists\footnote{If radiologists
have a shared tendency to perceive some artifact as a malignancy, the learned model should be able to assign large probability
to this region as well.}.

These probabilities are normalized for the case of GT created by a single annotator. If a GT mark is created by one annotator and the image is read
by the smallest allowed number of annotators $n^a_{min}$, the desired probability of malignancy is set to $P^\ast = k_d P^0$, where $P^0 =2/n^a_{max}$
is the smallest meaningful probability\footnote{probability, when only $2$ readers from the maximal number of them ($25$) mark the region as malignant}
and $k_d$ is a depression coefficient ($1/8$ was considered by us). In the case of the larger number of image readers
and GT created by a single annotator, its probability falls down proportionally to the number of readers $n^a$, so that
$P = \frac{P^\ast}{ n^a/n^a_{min}}$.


\section{MIL with soft labeling and $L_1$ regularization} \label{sec:softMIL}

In the MIL framework the data is considered to be aggregated into the so called bags
$\mathbf{x}_\mu, \mu=1 \ldots {\cal M}$, (${\cal M}$ is the number of bags). All the instances of the bag share the same extra bag-state label
being positive or negative. The bag-$\mu$ is considered to be negative if all its
instances $x_\mu^k,~k=1 \ldots S_\mu$ ($K_\mu$ is the number of samples in the bag $\mu$) are negative;
and positive if at least one its instance is positive.
The probability of the $\mu^{th}$ bag to be positive ($y_\mu=1$) is given by:
\begin{eqnarray}
p_\mu^{+}  = p(y_\mu=1 | \mathbf{x}_\mu) = 1 - \prod_{k=1}^{K_\mu}(1- P_{\mu k}^{+})\nonumber \\
P_{\mu k}^{+} =  \sigma(w^t x_{\mu k}), ~~~ \sigma(z) = \frac{1}{1+e^{-z}} \nonumber
\end{eqnarray}

We generalize the MIL concept to allow soft labeling, so that the instances are now aggregated into bags and each bag is malignant
with an apriori probability $\tilde{p}(y_\mu| \mathbf{x}_\mu)$. The standard ML (maximum likelihood) criterion learning objective function is replaced by the
minimization of the KL divergence between the probability distributions $\tilde{p}_\mu$ and
$p_\mu$, assuming the prior and estimated distributions being
i.i.d.: $D(w) =  \sum_{\mu=1}^M D^\mu_{KL}$, where:
\begin{eqnarray}
D^\mu_{KL}(\tilde{p}_\mu, p_\mu) = \tilde{p}_\mu^{+} log(\frac{\tilde{p}_\mu^{+}}{p_\mu^{+}}) +  (1-\tilde{p}_\mu^{+}) log(\frac{1-\tilde{p}_\mu^{+}}{1-p_\mu^{+}}) \nonumber
\end{eqnarray}
We minimize KL divergence with the $L_1$ constraint on the projection vector $w$ to prefer the parsimonious models.
\begin{eqnarray}
&& w_\ast = \arg \min_{w \in {\cal R}^d} {\cal F}(w) \nonumber \\
&& {\cal F}(w) = D(w) + \lambda \sum_{i=1}^d |w_i|, ~D(w) = \sum_{\mu=1}^M D_\mu(w) \nonumber \\
&& D_\mu(w) =  -[\tilde{p}_\mu^{+} log(p_\mu^{+}) +  (1-\tilde{p}_\mu^{+}) log(1-p_\mu^{+})] \label{Eq.KL.and.L1}
\end{eqnarray}

We use the conjugate gradient descend to find the model parameters and software developed by Schmidt et.al~\cite{Schmidt2007}.
In order to use the code the gradient and Hessian of the objective function have to be analytically calculated; these expressions are
evaluated and presented in the Appendix section~\ref{sec:Appendix}.

\subsection{Normalization issues}
Working with the regularization parameters it is worth to get some intuition about setting them reasonably.
In practice the data dimensionality (number of extracted features) changes between different versions.
This motivates us to normalize the $L_1$ norm by the input space dimensionality $d$ to:
\begin{eqnarray}
{\cal F}(w) =  D(w) + \lambda  \frac{1}{d} \sum_{i=1}^d |w_i|
\end{eqnarray}

Another issue is changing the number of data samples during training that naturally leads
to regularization parameter resetting. A simple idea is to normalize the KL-divergence term by the number of overall samples $M$:
\begin{eqnarray}
{\cal F}(w) = \frac{1}{M} D(w) + \lambda  \frac{1}{d} \sum_{i=1}^d |w_i|
\end{eqnarray}

The second one approach is more delicate as also tries to resolve imbalance problem, by normalizing separately per
soft positive and hard negative KL-divergence term.
\begin{eqnarray}
{\cal F}(w)  = \frac{1}{M^{+}} D^{+}(w) + \frac{1}{M^{-}} D^{-}(w) +  \lambda  \frac{1}{d} \sum_{i=1}^d |w_i|
\end{eqnarray}

From the computational view point any of these normalizations is equivalent to a per bag-normalization.

\section{Lesion Detection in CXR CAD}
\subsection{CAD objective and the experimental design.} \label{sec:experimental design}
In the computer aided diagnosis (CAD) the goal is usually defined as detection of malignant regions in the digital images.
However, in our statement, there is no any additional information about the
malignancy state of the radiologist marks, such as biopsy results or
observation from other modalities such as CT, for example. In the absence
of the golden ground truth we resume to a standard practice used by radiologists.
Namely, if two readers are consistent in considering the region as being malignant;
our system should be able to detect this.

In other words,
the goal of our system should be to assist the radiologists in making diagnosis so that
they are satisfied with the system, rather than detecting unknown ground truth that may
be also invisible to the eye in the X-ray images.
In summary, while the soft classifier learns the probability of the region to be marked
as malignant by radiologists, the final ROC results~\cite{Fawcett06} reporting the sensitivity and number of false positives
in an image are presented as if the pseudo golden GT corresponds to GTs marked by more than one reader.

The soft classifier proposed by us is compared with two other binary classifiers:
\begin{itemize}
\item {\bf B1}: the binary RVMMIL classifier discriminating the pseudo golden GTs.
\item {\bf B2}: the mixture of binary RVMMIL classifiers trained per the experts. The mixture classifier is an extra binary MIL
logistic regression network with the input being the expert classifier continuous score outputs. The final logistic network goal is to discriminate
a binary pseudo golden GTs.
\end{itemize}
The goal of the binary classifiers  {\bf B1-B2} is to label the suspicious candidates as
being malignant from the radiologists point of view\footnote{marked more than once by different radiologists in our case}.
In other words, it has to reduce the
number of false positives (FPs) without decrease in the sensitivity.
In binary classification, the candidates that are close to the pseudo golden GT mark are arranged into a positive bag;
all the other candidates are considered as negative.

The CXRCAD system consists of the three steps: (1) candidate generation, (2) feature extraction and finally (3) classification.
In the first step the suspicious regions (lesion candidates) are found.
While this step detects a lot of the anomalies, the number of extracted false positives is extremely high.
In order to reduce the number of false positives, features describing the suspicious regions are extracted and
classification is performed in this feature space.
\subsection{Data Description}
The suspicious region features are presented by two groups: a relatively
small set of specific features and a huge number of standard intensity based and texture
features (see \cite{Wei97,Haralick73}). The specific features are
elaborately extracted to reflect the heuristic nature of the
lesions, such as spiculation features, for example
\cite{SampatBovik06} or the region label, indicating if it falls in the rib region.
The number of specific features is about
$150$ and the number of texture-like is about $1000$.

\begin{figure}[htb]
\begin{center}
\begin{tabular}{cc}
a & b \\
\includegraphics[height=4cm,width=3cm]{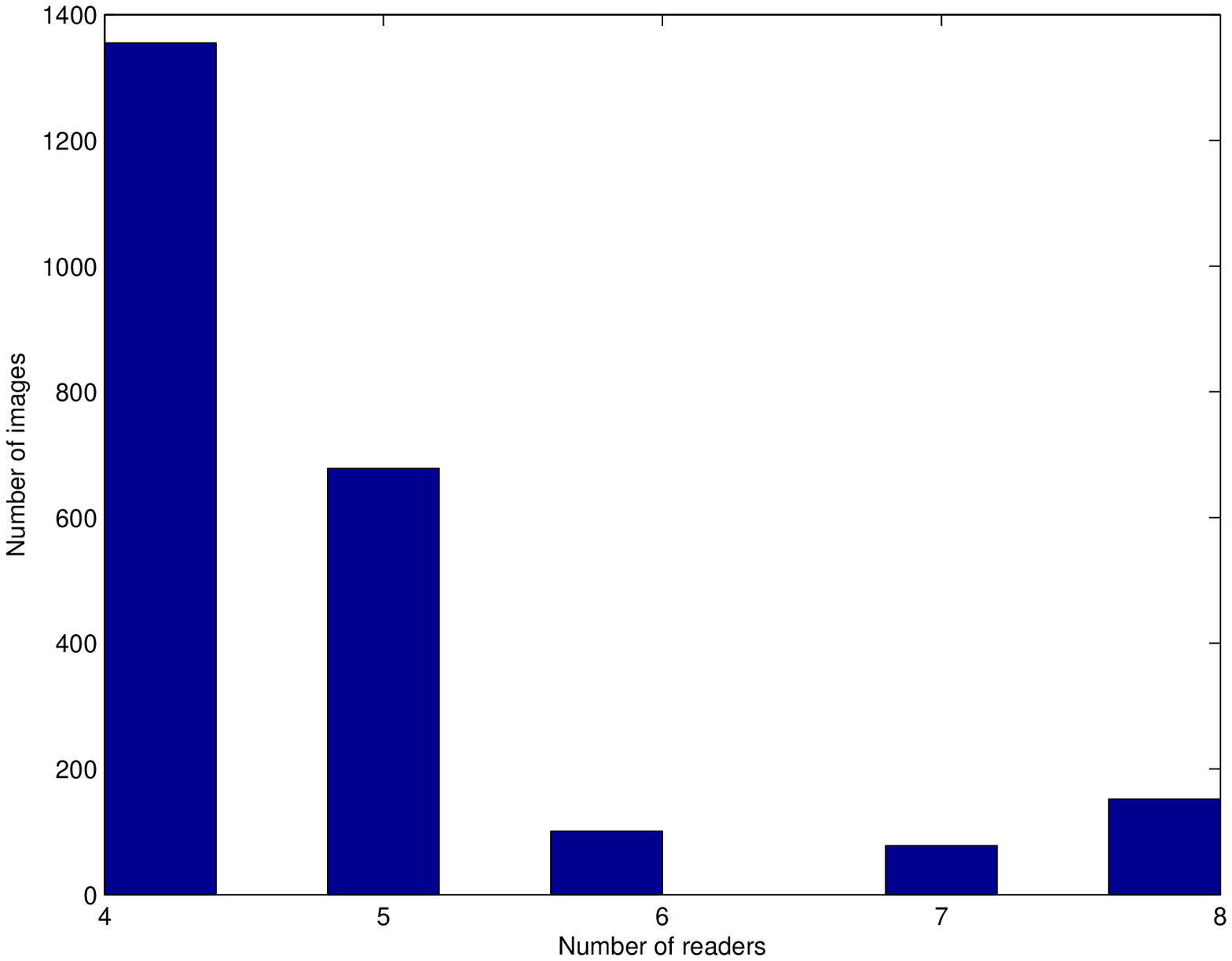} & \includegraphics[height=4cm,width=5cm]{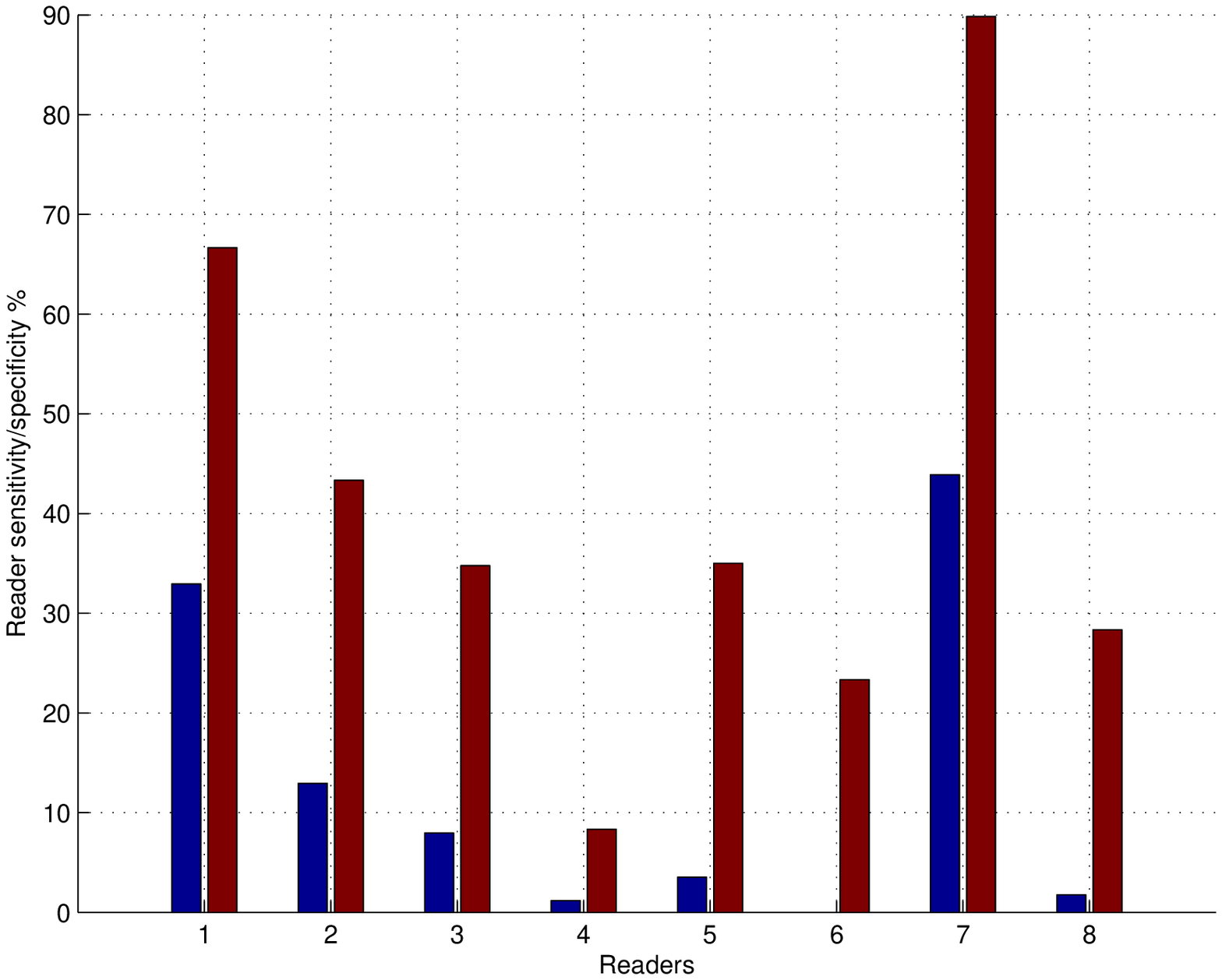}  \\
\end{tabular}
\caption{{\sf Annotation Information.} a. The annotators histograms, i.e. number of images read by a certain number of annotators.
b. The sensitivity and false positive rates of the readers in respect to the pseudo golden GTs shown by red and blue bars, respectively.}
\label{fig:readerStats}
\end{center}
\end{figure}

The number of overall images at our disposal is $1978$, among them the number of potentially malignant images having at least one GT mark
is equal to $1072$. The number of overall marks by all the experts is equal to $2364$ and the number of generated pseudo golden GTs are
equal to $322$. The $8$ readers participated in the annotation process. The annotator histogram, i.e. the number of images read by a certain number of annotators is presented in Figure~\ref{fig:readerStats}a.
As can be seen most of the images were marked by $4-5$ annotators and only a small number by more than $5$.
The sensitivity and false positive rates of the readers
in respect to the pseudo golden GTs are presented in Figure~\ref{fig:readerStats}b.

\section{Results}
The data is divided into three subsets of the same size: training, validation and test.
The "training + validation" sets are used for training models and setting the model parameters
and results are reported on the unseen test data. We consider three models $Soft, B1, B2$ and
compare their performance and complexity.

\subsection{MIL with soft labeling and $L_1$ regularization.}
\begin{figure}[htb]
\begin{center}
\begin{tabular}{c}
a. $FP=0.5$ \\
\includegraphics[height=5cm,width=7.5cm]{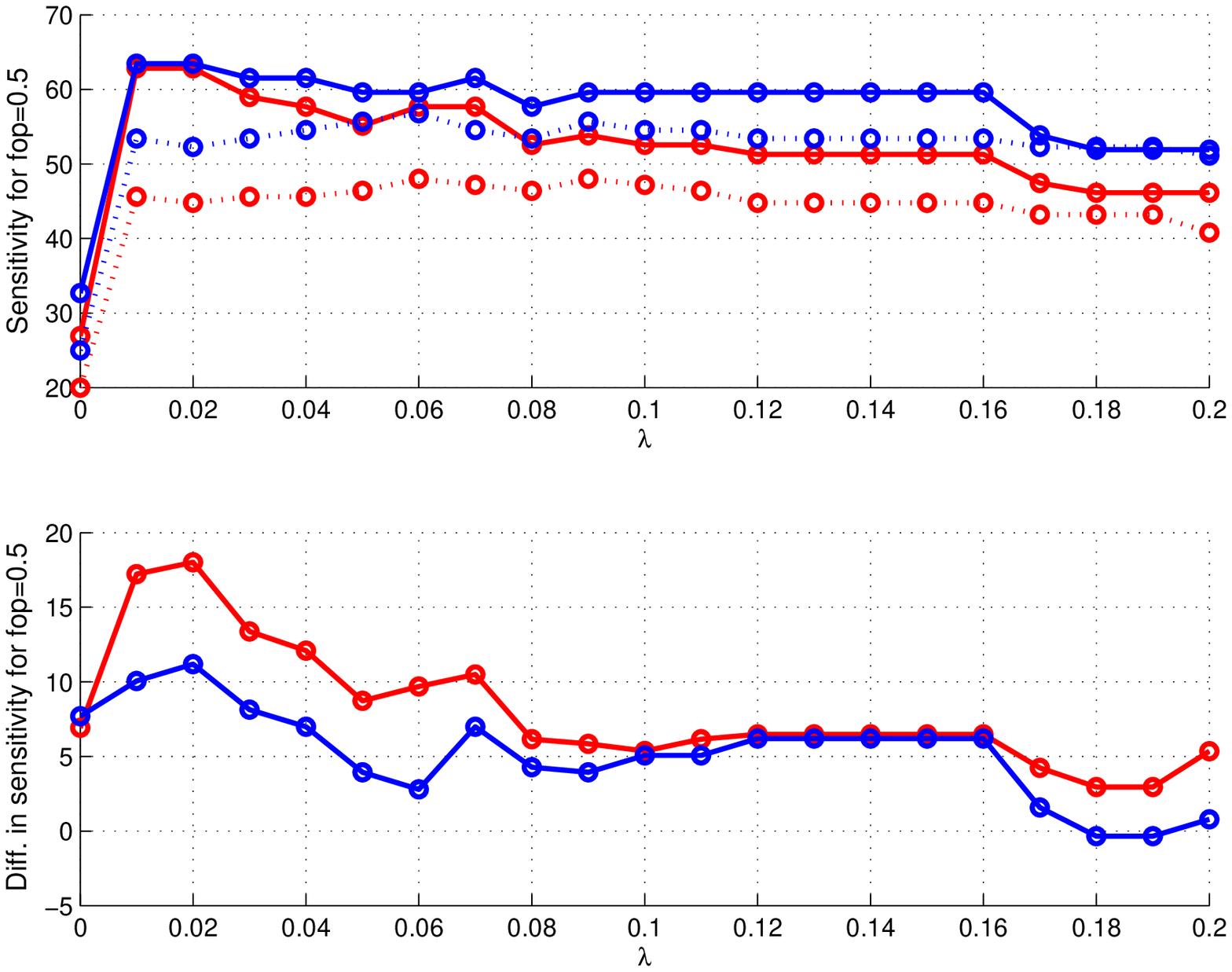} \\
b. $FP=1.0$ \\
\includegraphics[height=5cm,width=7.5cm]{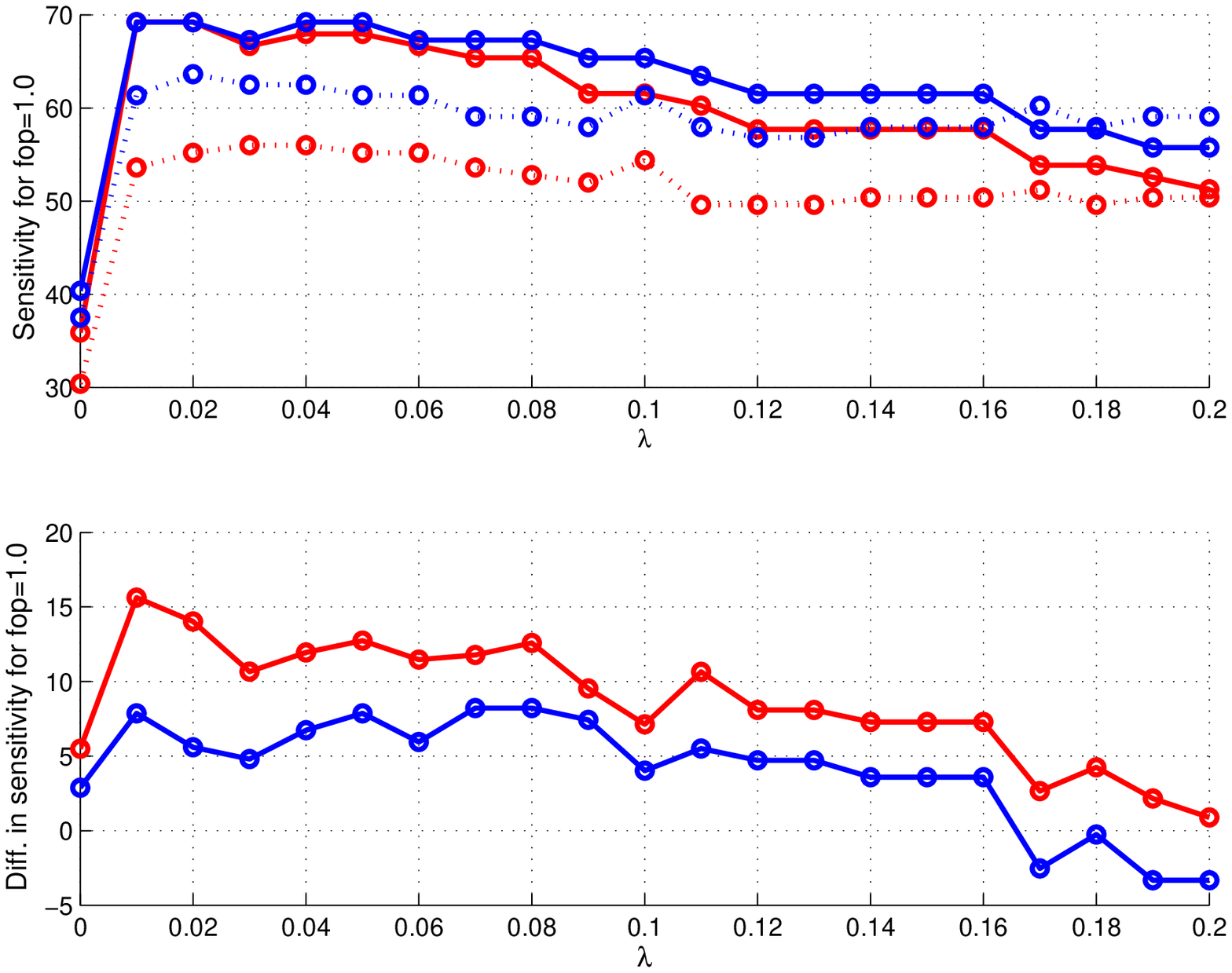} \\
\end{tabular}
\caption{{\sf Selection of the optimal regularization parameter.} The top graphs in sufigures a. and b. show sensitivity for different values of
regularization parameters on the "training + validation" sets, for FP values $0.5$ and $1.0$, respectively.
The curve marked by blue stands for image sensitivity and the red one for GT sensitivity. The bold and dashed lines are used to show results on "training + validation" sets, respectively.
The low plots show the difference in sensitivity on the validation and training sets.
The optimal selected $\lambda=0.06$ leads to the largest sensitivity for image and GT sensitivity and the relatively small absolute difference in sensitivity
between "training + validation" sets for both FP values.}
\label{fig:optimalLambda}
\end{center}
\end{figure}

The first vital system requirement is finding less than $0.5$ false positives (FP) per image. The higher FP values are not acceptable by radiologists
because it complicates their work, leading to tiredness and errors. The second requirement is a high sensitivity both on GTs and images\footnote{When measuring image
sensitivity, it is sufficient to detect at least one GT in the image. In other words, finding of at least one real nodule by CAD is considered as a success.}.
The sensitivity of the classifiers for the $FP=0.5$
per different regularization $\lambda$ parameters are presented in Figure~\ref{fig:optimalLambda}. As can be seen (Figure~\ref{fig:optimalLambda}.a) for $FP=0.05$
there exists an interval of regularization values $\lambda \in [0.05~0.16]$ where regularized soft models have large sensitivity values and relatively small difference in sensitivity between
"training + validation" sets. In order to shrink this interval and select the optimal $\lambda$ value, the models sensitivities for larger $FP=1.0$ are considered as well
(Figure~\ref{fig:optimalLambda}.b). The optimal interval for $FP=1.0$ lies between  $\lambda \in [0.03~0.08]$. The final optimal value was selected by us in the cross section of
the optimal intervals for $FP=0.5$ and $FP=1.0$; and was set to $\lambda^\ast = 0.06$.

The final soft classifier is obtained by retraining the soft classifier for the selected $\lambda^\ast = 0.06$ and
the results are reported on the unseen test data, that was not used in any step of the soft classifier
training\footnote{The reported results do not take into account possible failure in candidate generation. We report and compare classification models only.}.
The ROCs for for the final soft classifier are presented in Figure~\ref{fig:performance}, red curves.
We also note that the FP slightly moves forward to 0.52 on the test data and there is slight decrease in the sensitivity.

\subsection{Comparison with the binary classifiers}
In this section the optimally regularized soft model is compared with two binary classifiers $B1$ and $B2$ (see Section~\ref{sec:experimental design}).
The RVM MIL classifier  ($B1$ model) is trained on the merged "training + validation" data sets and
tested on the unseen test data.

Though, the complexity of the RVM MIL model\footnote{number of selected features} is defined automatically during learning via the hyper parameters;
our multiple experiments with the RVM MIL show that it is not as trivial as appear in~\cite{RaykarlEtal08}.
The final RVM MIL model complexity and performance results are dependent on the other
control parameters. Ideally, the setting of these control parameters should be performed
on the validation data set. Instead, we favor the RVM MIL by training it straightforward on merged  "training + validation" data sets and selecting parameters leading
to the best results. The mixture of RVM MIL experts models ($B2$ model) is also trained on the merged "training + validation" data sets.
The final scores are mixed by an extra binary network.

\begin{figure}[htb]
\begin{center}
\begin{tabular}{c}
GT sensitivity \\
\includegraphics[height=6cm,width=8.0cm]{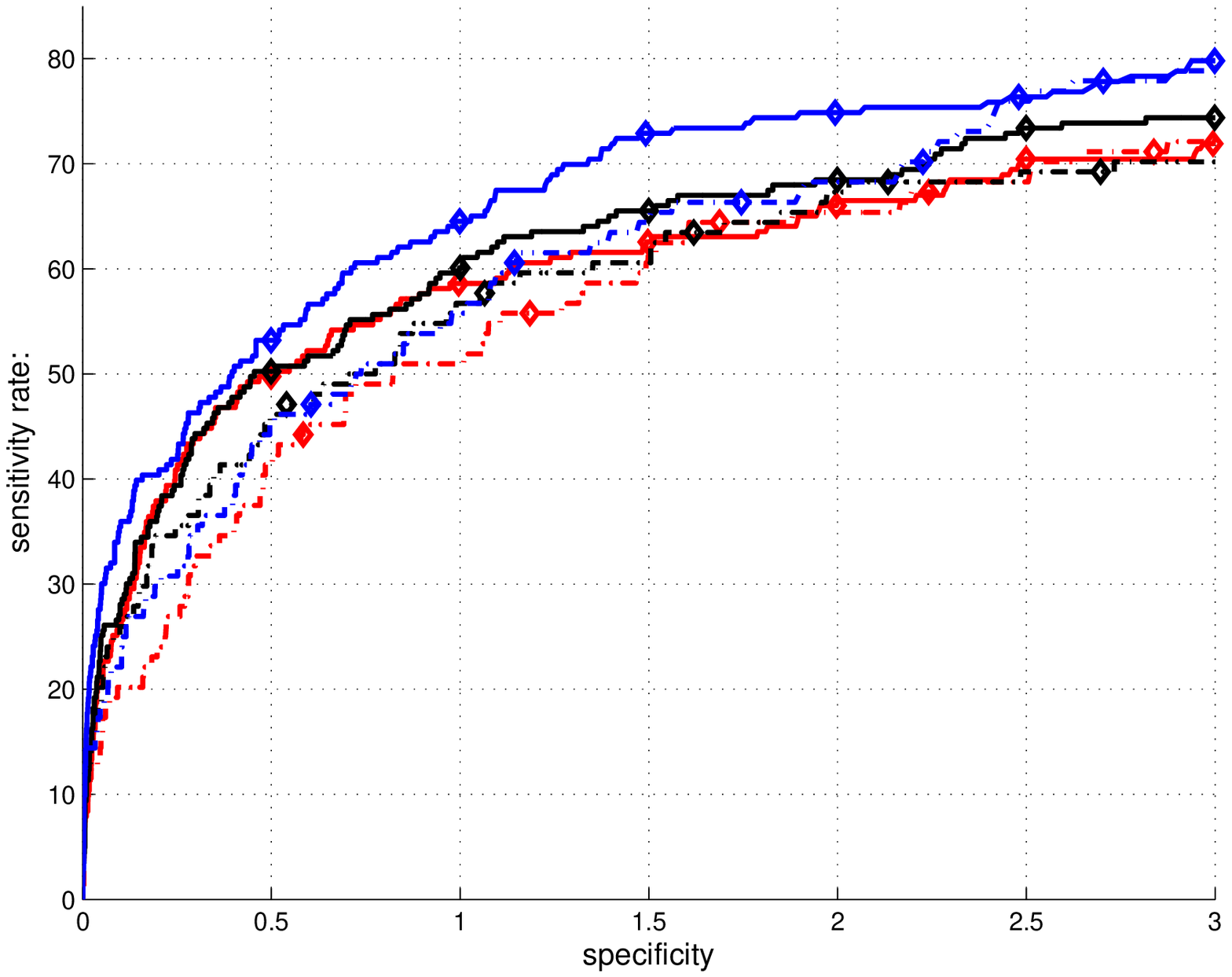} \\
Image Sensitivity \\
\includegraphics[height=6cm,width=8.0cm]{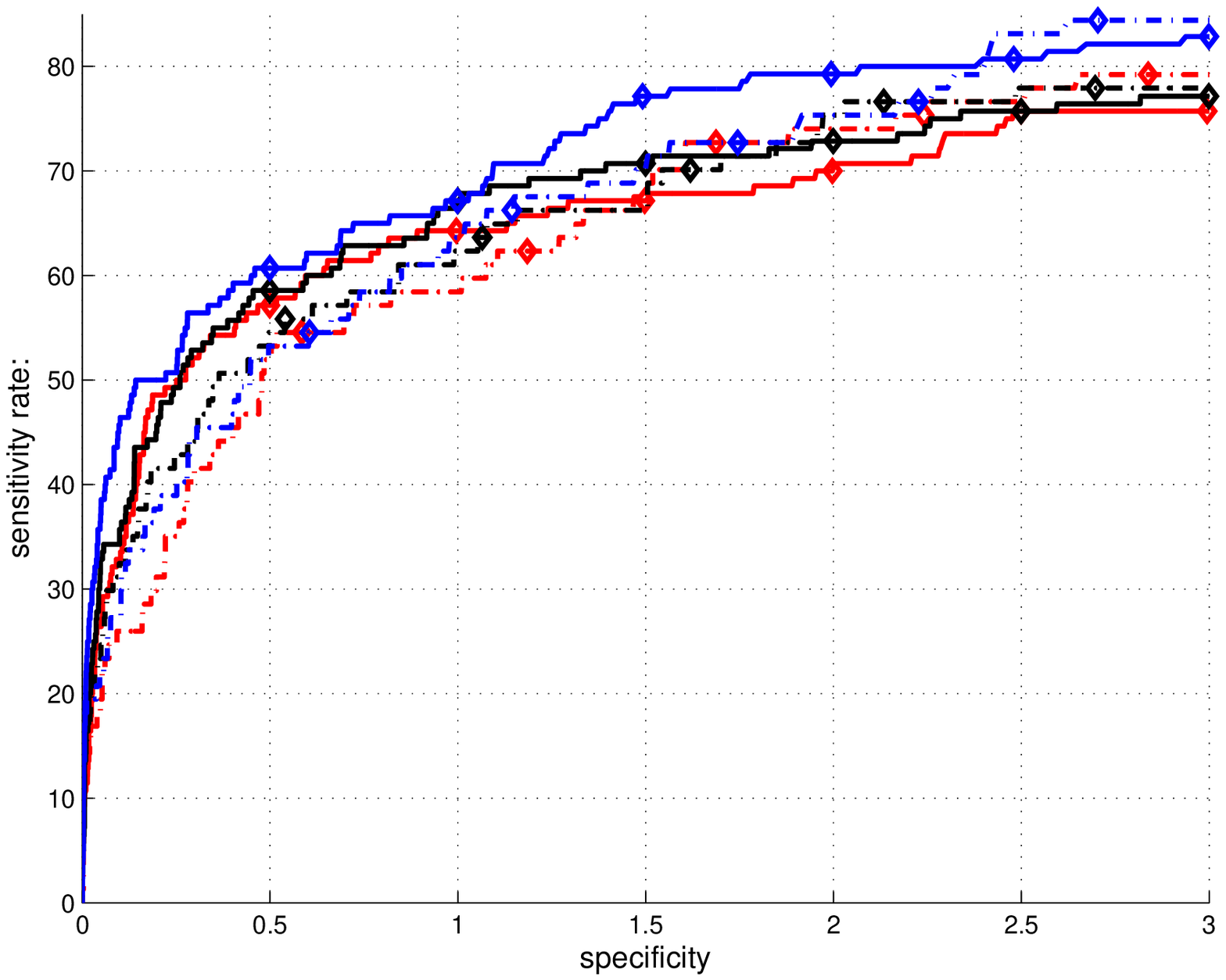} \\
\end{tabular}
\caption{{\sf The models ROC results.} The three models: $Soft$, $B1$ and $B2$ are marked by red, black and blue colors, respectively.
The dashed lines show results on the test data and bold on the merged  "training+validation" data. The sensitivity values at $FP$ values $0.5:0.5:3$
are marked by diamond.}
\label{fig:performance}
\end{center}
\end{figure}

The results of all three models on "training+validation" and test data are presented in Figure~\ref{fig:performance} and Table~\ref{tab:performance}.
\begin{table}[h]
\begin{center}
\begin{tabular}{|l|l|l|l|l|l|l|} \hline 
\multicolumn{7}{|c|} {{\bf Soft Model: lambda=0.060000} Number selected features 67} \\
\hline
\multicolumn{7}{|l|}{\sf "training+validation" data} \\ \hline 
&            \multicolumn{6}{|c|}{ FP per image} \\ \hline
				  &  0.50 &  1.00 &  1.50 &  2.00 &  2.50 &  3.0 \\
GT sensitivity    & 49.75 & 58.62 & 62.56 & 66.01 & 70.44 & 71.92 \\
Image sensitivity & 57.14 & 64.29 & 67.14 & 70.00 & 75.71 & 75.71 \\
\hline
\multicolumn{7}{|l|}{\sf test data}  \\ \hline 
& \multicolumn{6}{|c|}{induced  FP per image} \\ \hline
				  &  0.58 &  1.19 &  1.69 &  2.24 &  2.84 &  3.37 \\
GT sensitivity    & 44.23 & 55.77 & 64.42 & 67.31 & 71.15 & 75.96 \\
Image sensitivity & 54.55 & 62.34 & 72.73 & 75.32 & 79.22 & 80.52 \\
\hline
\multicolumn{7}{|c|} {{\bf Model B1:} Number selected features 88} \\
\hline
\multicolumn{7}{|l|}{\sf "training+validation" data} \\ \hline
&            \multicolumn{6}{|c|}{ FP per image} \\ \hline 
				  &  0.50 &  1.00 &  1.50 &  2.00 &  2.50 &  3.00 \\
GT sensitivity    & 50.25 & 60.10 & 65.52 & 68.47 & 73.40 & 74.38 \\
Image sensitivity & 58.57 & 67.14 & 70.71 & 72.86 & 75.71 & 77.14 \\
\hline
\multicolumn{7}{|l|}{\sf test data}  \\ \hline 
& \multicolumn{6}{|c|}{induced  FP per image} \\ \hline
                  &  0.54 &  1.07 &  1.62 &  2.13 &  2.70 &  3.17 \\
GT sensitivity    & 47.12 & 57.69 & 63.46 & 68.27 & 69.23 & 70.19 \\
Image sensitivity & 55.84 & 63.64 & 70.13 & 76.62 & 77.92 & 77.92 \\
\hline
\multicolumn{7}{|c|} {{\bf Model B2:}  Number selected features 203} \\
\hline
\multicolumn{7}{|l|}{\sf "training+validation" data} \\ \hline 
& \multicolumn{6}{|c|}{induced  FP per image} \\ \hline
                  &  0.50 &  1.00 &  1.49 &  1.99 &  2.48 &  3.00 \\
GT sensitivity    & 53.20 & 64.53 & 72.91 & 74.88 & 76.35 & 79.80 \\
Image sensitivity & 60.71 & 67.14 & 77.14 & 79.29 & 80.71 & 82.86 \\
\hline
\multicolumn{7}{|l|}{\sf test data}  \\ \hline 
& \multicolumn{6}{|c|}{induced  FP per image} \\ \hline
				  &  0.61 &  1.14 &  1.75 &  2.23 &  2.70 &  3.28 \\
GT sensitivity    & 47.12 & 60.58 & 66.35 & 70.19 & 77.88 & 78.85 \\
Image sensitivity & 54.55 & 66.23 & 72.73 & 76.62 & 84.42 & 84.42 \\
\hline
\end{tabular} 
\caption{ ROCs characteristics: {\sf The three models sensitivities versus different FP (false positive) candidates per image.}}
\label{tab:performance}
\end{center}
\end{table}

As can be seen all three models achieve almost the same performance, however the $Soft$ model is the most
compact as uses only $67$ features, comparing with $B1$ and $B2$ models that use $88$ and $203$ features, respectively.
The nice property of the regularized soft model is an easily perceivable control on the model complexity; as regularization parameter $\lambda$
increases the number of selected features decreases that should finally lead to more robust results.
As can be seen the FP value is slightly moving forward on the test data for all three models.

\begin{figure}[htb]
\begin{center}
\begin{tabular}{c}
\includegraphics[height=6cm,width=8.0cm]{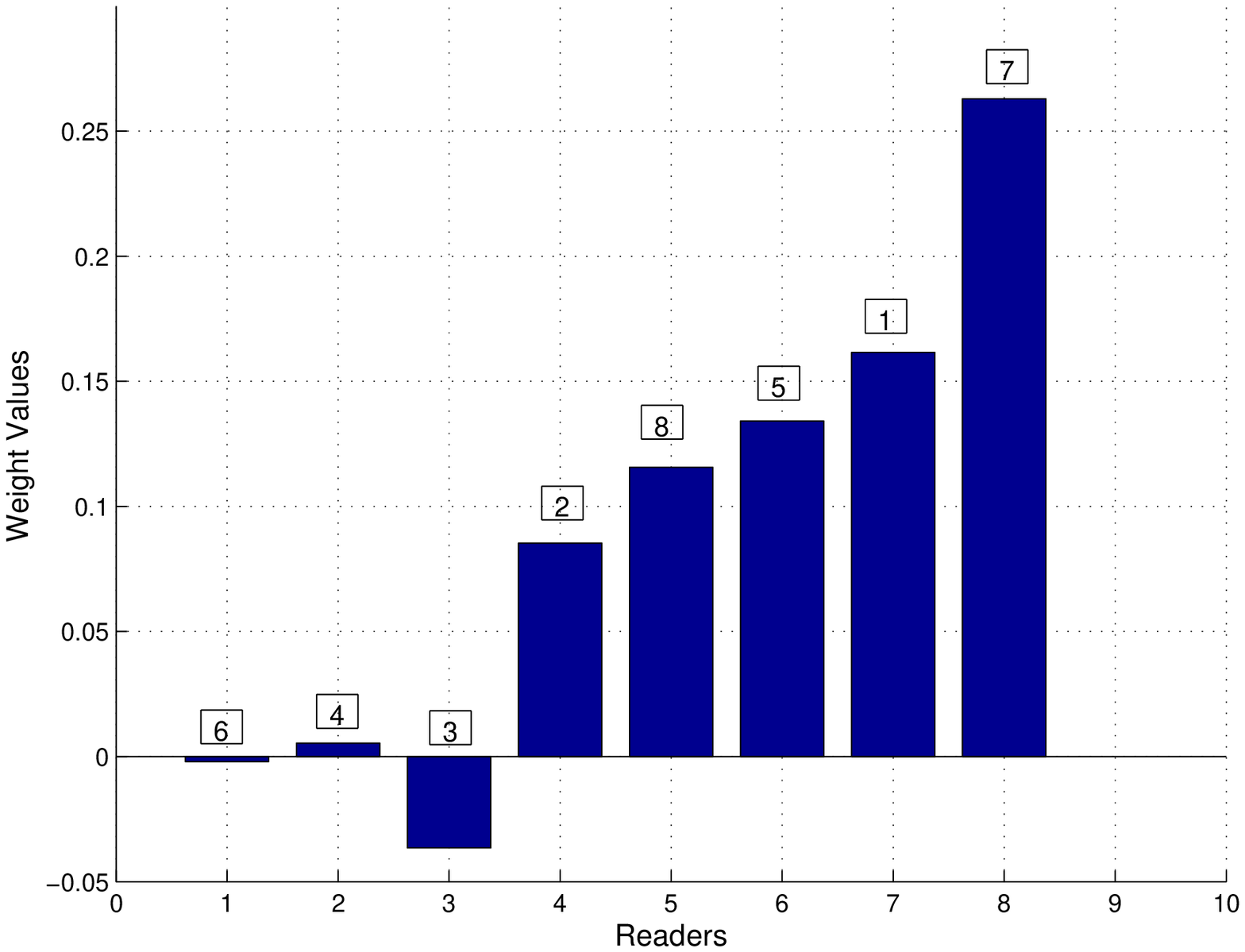} \\
\end{tabular}
\caption{{\sf Reader Importance.}  The weights (mixing coefficients) of the mixture network. The numbers in boxes indicate the indices of the readers
in correspondence with Figure~\ref{fig:readerStats}b. Note that the readers $7$ and $1$ got the largest weights and these readers have also the largest sensitivities
(Figure~\ref{fig:readerStats}b, red columns). }
\label{fig:readerImportance}
\end{center}
\end{figure}

Though learning the $B2$ model is extensive, it provides also some estimation of the readers importance.
The final weights of the mixture network is presented in Figure~\ref{fig:readerImportance}.

\pagebreak

\section*{Discussion}
In this paper we investigated how multiple annotation can be used for training the CAD system.
We introduced the notion of pseudo golden GT and considered three learning models to address the problem.
One of the models introduced by us is a soft classifier that models the desired probabilities of the regions
to be marked by the radiologists. The usage of the soft classifier is a good alternative to using the RVM MIL
classifier or mixture of RVM MIL experts model that enables additionally to model the desired probabilities.
This model is the most compact between the three learned models,
that is desirable in practice from the computational and explorative view point.

It is interesting to compare the introduced models in different scenarios with the growing number of images and
number of annotators, when the number of images per annotators is very small. We are planning to conduct
such a research in the future as such data will become available\footnote{Marking the CXR CAD images is quite an expensive process,
requiring the financial investment.}.

\section{Appendix}\label{sec:Appendix}

In the case of the hard apriori labels  $\tilde{p}_\mu^{+}=1$, we get exactly the same objective function as in \cite{RaykarlEtal08},(Eq.8), that is the maximum likelihood of the data
under the i.i.d. sample assumption:
\begin{eqnarray}
&& w_\ast = \arg \max_{w \in {\cal R}^d} L(w),
~~~ L(w) = \sum_{\mu=1}^M L_\mu \nonumber \\
&& L_\mu = y_\mu log(p_\mu^{+}) +  (1-y_\mu) log(1-p_\mu^{+}).  \label{Eq.MIL.Logistic}
\end{eqnarray}

Moreover, comparing two objective functions allows to recalculate gradient and Hessian of the $D(w)$ very easily, by observing that
\begin{eqnarray}
D_\mu(w)= -[ \tilde{p}_\mu^{+} L_{\mu|y_\mu=1} + (1-\tilde{p}_\mu^{+}) L_{\mu|y_\mu=0}]\label{Eq.Comparing}
\end{eqnarray}

Below we present the gradient and Hessian expressions as appear in paper \cite{RaykarlEtal08}, in a bit more compact way:
\begin{eqnarray}
&& g(w) = \nabla_w L(w) = \sum_{\mu} g_\mu(w), ~~~ g_\mu(w) = \nabla_w L_\mu(w) \nonumber \\
&& g_\mu(w) = [y_\mu \beta_\mu-(1-y_\mu)] x_\mu^g, \nonumber \\
&& \boxed{x_\mu^g=\sum_{j=1}^{K_\mu} x_{\mu j} \sigma(w^t x_{\mu j}) = \sum_{j=1}^{K_\mu} x_{\mu j} P^{+}_{\mu j}} \label{Eq.MIL.Grad} \\
&& H(w) = \sum_{\mu} H_{\mu}(w) \nonumber \\
&& H_{\mu}(w) = [y_\mu \beta_\mu-(1-y_\mu)] \times \nonumber \\
&& \sum_{j=1}^{K_\mu} x_{\mu j} x_{\mu j}^t c_{\mu j} -
y_\mu \beta_{\mu}(\beta_{\mu}+1) x_\mu^g (x_\mu^g)^t \label{Eq.MIL.Hes} \\
&& c_{\mu j} =  \sigma(w^t x_{\mu j}) \sigma(-w^t x_{\mu j}), ~~~ \beta_\mu = \frac{1-p_\mu^{+}}{p_\mu^{+}} = \frac{1}{p_\mu^{+}}-1 \nonumber
\end{eqnarray}

Since gradient and Hessian are linear operators and $D(w)$ is a linear function of $L_\mu$, it is easily seen that:
\begin{eqnarray}
&& \boxed{g^{new}(w) = \nabla_w D(w) = \sum_{\mu} g^{new}_\mu(w)} \nonumber \\
&& g^{new}_\mu(w) = \nabla_w D_\mu(w) \nonumber \\
&& g^{new}_\mu(w) = -[\tilde{p_\mu}^{+} g_{\mu|y_\mu=1}(w) + (1-\tilde{p_\mu}^{+})g_{\mu|y_\mu=0}(w)]\nonumber \\
&& \boxed{H^{new}(w) = \sum_{\mu} H^{new}_{\mu}(w)} \nonumber \\
&& H^{new}_{\mu}(w) = -[\tilde{p_\mu}^{+} H_{\mu|y_\mu=1}(w) + (1-\tilde{p_\mu}^{+})H_{\mu|y_\mu=0}(w)]\nonumber
\end{eqnarray}

Noticing that: $[y_\mu \beta_\mu-(1-y_\mu)]|_{y_\mu=1} = \beta_\mu$ and $[y_\mu \beta_\mu-(1-y_\mu)]|_{y_\mu=0} = -1$, we can simplify $g^{new}_\mu(w)$ to
\begin{eqnarray}
&&g^{new}_\mu(w) = -[(\tilde{p_\mu}^{+}\beta_\mu - (1-\tilde{p_\mu}^{+}))x_\mu^g]\nonumber \\
&& = -[(\tilde{p_\mu}^{+}(\beta_\mu+1) - 1)x_\mu^g] = -[(\frac{\tilde{p_\mu}^{+}}{p_\mu^{+}}-1)x_\mu^g] \nonumber \\
&&\boxed{g^{new}_\mu(w) =  -[(\frac{\tilde{p_\mu}^{+}}{p_\mu^{+}}-1)x_\mu^g]}
\end{eqnarray}
and Hessian per bag to:
\begin{eqnarray}
&& H^{new}_\mu(w) = -[(\frac{\tilde{p_\mu}^{+}}{p_\mu^{+}}-1) \times \nonumber \\
&& (\sum_{j=1}^{K_\mu} x_{\mu j} x_{\mu j}^t c_{\mu j}) -
\tilde{p_\mu}^{+}\beta_{\mu}(\beta_{\mu}+1) x_\mu^g (x_\mu^g)^t]
\end{eqnarray}

In the code implementation the next expressions are useful:
\begin{eqnarray}
&&\beta_{\mu}(\beta_{\mu}+1) = (\frac{1}{p_\mu^{+}}-1}){\frac{1}{p_\mu^{+}} = \frac{1-p_\mu^{+}}{(p_\mu^{+})^2} \nonumber \\
&&\boxed{\beta_{\mu}(\beta_{\mu}+1) = \frac{1-p_\mu^{+}}{(p_\mu^{+})^2} }\nonumber
\end{eqnarray}

We note also that $\sigma(z_j)$ is associated with the probability $P_j^{+}$ of the instance $j$ (a linear index independent of the bag) to be positive.
This enables us to simplify some other entities:
as $\sigma(z_j) = \frac{1}{1+exp(-z_j)} \Rightarrow \sigma(-z_j)=\frac{1}{1+exp(z_j)} = \frac{exp(-z_j)}{1+exp(-z_j)} = P_j^{+} (\frac{1}{P_j^{+}}-1) = 1 - P_j^{+}$
and finally
\begin{eqnarray}
\boxed{\sigma(z_j)\sigma(-z_j) = P_j^{+}(1-P_j^{+}).}
\end{eqnarray}

In reality, we deal with the imbalanced problem with the mixed hard and soft labels, where we have a large number of hard negative labels.
While there is not any problem to use the same expressions for hard labels as for soft ones in implementation it is faster to consider the
gradient $g^{new, -}_\mu$ and Hessian in a hard negative case $H^{new, -}_\mu$, separately:
\begin{eqnarray}
\boxed{g^{new, -}_\mu(w) = -[-x_{\mu 1}^t P_{\mu j}^{+}]} \nonumber \\
\boxed{H^{new, -}_\mu(w) = -[-x_{\mu 1} x_{\mu 1}^t c_{\mu j}]}
\end{eqnarray}
where $x_{\mu 1}$ a single instance in the $\mu$-th bag.

\subsection{Different number of annotators}
One of the ways to estimate the apriori probability of the bag is considering the proportion of the bag to be marked by different readers
(annotators) to their overall number. In practice the number of readers (annotators) may be different for different images;
this is not reflected in our formulation. Though we can artificially set the prior probabilities to be dependent on the number
of readers (by treating the probabilities to be a measure of our belief of the bag to be positive, rather than the probability of the bag being positive);
the more natural treatment is described below.

Lets assume that the same image is read by $N=2n$ readers and $K=2k$ of them marked the same
object (bag) as a positive one. The estimated a-priori probability of the bag to be positive is  $\tilde{p}^{+} = k/n$. Let us divide evenly the readers
into two independent annotator groups of size $n$, from which $k$ marked the object as the positive. The estimated
probability of the bag to be positive in each of the group remains the same $\tilde{p}^{+}$. Assume the groups are independent and consider their
work independently\footnote{like having two independent images, marked by each group separately. Here we decline from the i.i.d. data assumption;
but anyway this assumption is rarely has place in practice; though the i.i.d. models lead to good results}.
In this case, we can think about the $\mu$-th bag term in the optimization function as being repeated twice in Eq.~\ref{Eq.KL.and.L1} as $2DL(\tilde{p}^{+},p)$.
In comparison, if only $n$ annotators participate in the marking process the same term appears only once in Eq.~\ref{Eq.KL.and.L1} as $DL(\tilde{p}^{+},p)$.

We generalize this observation above to modify the optimization function to take into account different number of annotators per image as:
\begin{eqnarray}
&& {\cal F}(w) = D^m(w) + ||w||_{L_1} \nonumber \\
&& D^m(w)=\sum_{\mu=1}^M a_\mu D_\mu(w), ~~~ a_\mu = \frac{n^a_\mu}{n^a_{max}},
\end{eqnarray}
where $n^a_\mu$ are the number of annotators per $\mu$-bag and $n^a_{max}$ is the maximal number of annotators.
The gradient and Hessian expressions will be modified as:
\begin{eqnarray}
&& g^{new}(w) = \sum_{\mu} a_\mu g^{new}_\mu(w) \nonumber \\
&& H^{new}(w) = \sum_{\mu} a_\mu   H^{new}_{\mu}(w) \label{Eq.modif}
\end{eqnarray}

\section*{Acknowledgments}
The authors would like to thank David Leib for his advice and help in the preparation of the paper.


\end{document}